# Methodology for Detection of QRS Pattern using Secondary Wavelets

T.R.Gopalakrishnan Nair
ARAMCO Endowed chair-Technology
PMU,KSA
Vice President, RIIC
Dayananda Sagar Institutions
Bangalore, India
trgnair@ieee.org

Geetha A P
Advanced Signal & Image Processing
Group, RIIC
Dayananda Sagar Institutions
Bangalore, India
apgeetha@yahoo.com

Asharani
Dept.of E&C
JNTUHEC
Hyderabad,India
ashajntu1@yahoo.com

*Abstract—Applications of wavelet transform to the field of health care signals have paved the way for implementing revolutionary approaches in detecting the presence of certain abnormalities in human health patterns. There were extensive studies carried out using primary wavelets in various signals like Electrocardiogram (ECG), sonogram etc. with a certain amount of success. On the other hand analysis using secondary wavelets which inherits the characteristics of a set of variations available in signals like ECG can be a promise to detect diseases with ease. Here a method to create a generalized adapted wavelet is presented which contains the information of QRS pattern collected from an anomaly sample space. The method has been tested and found to be successful in locating the position of R peak in noise embedded ECG signal.*

## I. Introduction

Wavelet transform is an efficient mathematical tool for local analysis of nonstationary and transient signals. Application of wavelet transform to estimate, characterize and identify properties of signals has been successful in many fields like astronomy, nuclear engineering, signal and image processing and fractals.

The wavelet analysis procedure is to adopt a wavelet prototype function, called an analyzing wavelet or mother wavelet function. A large section of localized waveforms can be employed as wavelet provided that they satisfy predefined mathematical criteria. The continuous wavelet transform (CWT) of a signal x(t) is defined as

$$W(a,b) = \frac{1}{\sqrt{a}} \int_{-\infty}^{\infty} x(t) \Psi^* \left(\frac{t-b}{a}\right) dt \quad (1)$$

where $\Psi^*(t)$ is the complex conjugate of the analyzing wavelet function ψ(t), parameters a and b is the dilation and location parameter of the wavelet respectively. However for a function to be classified as a wavelet, it has to satisfy certain mathematical criteria such as

i) It has to posses finite energy
$$E = \int_{-\infty}^{\infty} |\psi(t)|^2 dt < \infty \quad (2)$$

ii) If $\widehat{\Psi}(f)$ is the Fourier transform of ψ (t), i.e.

$$\psi^*(f) = \int_{-\infty}^{\infty} \psi(t) e^{-j(2\pi ft)} dt \quad (3)$$

then the necessary condition is

$$C_g = \int_0^{\infty} \frac{|\widehat{\psi}(f)|^2}{f} df < \infty \quad (4)$$

Criterion (ii) implies that the wavelet has no zero frequency components. This criterion is known as admissibility condition and $C_g$ is the admissibility constant [1].

Wavelet transform can be considered as the projection of a signal x(t) into a family of functions that are the normalized, dilated and translated versions of a prototype function (the wavelet). Wavelet transforms can comprise an infinite set of possible basis functions. Primary wavelets or the adapted wavelets are used as basis functions. There are many primary wavelet bases designed for general applications such as Meyer, Mexican hat, Haar, Daubechies etc [1].

This paper provides an analysis of ECG signal in order to accurately identify QRS complex peak. ECG records the electrical activity of heart over a period of time as detected by electrodes attached to the outer surface of the skin and recorded by a device external to the body. Initial part of the waveform - P wave, represents activity of the atria. The following part is the QRS complex which is produced by the contraction of ventricles. The subsequent return of the ventricular mass to a rest state - re-polarization produces the T wave. R wave detectors are extremely useful tools for the analysis of ECG signals. Making an algorithm for detection of R-peak in QRS complex is a difficult task. This is due to the variability of ECG patterns caused by noise and artifacts [2].

Since the quality of performance of an automated ECG analysis system depends highly on the reliable detection of R-peaks, the aim of this research work is to locate R peaks accurately. Heart rate variability measures rely heavily on the accuracy of the QRS feature detection on the digitized ECG signal [3]. QRS detection is difficult due to the physiological variability of the QRS complex. Other ECG components such as the P or T wave which look similar to QRS complexes often lead to wrong detections. Noise that is embedded within the ECG signal makes the detection of R-peak more complex. Noise sources include muscle noise, artifacts due to electrode motion, power-line interference and baseline wander.

Analyzing an ECG waveform using pattern adapted wavelets gives a better method for accurate analysis because of the rhythmic nature of the signal. Here, we have considered a set of signal patterns for adapted wavelet formation by considering typical variations in QRS complex. These adapted wavelets are used to bring out an accurate and easy method for the detection of R-peak in an ECG signal.

## II. RESEARCH BACKGROUND

An adapted wavelet is the best basis function designed for a given signal representation. There are several methods to construct pattern-adapted wavelets. Authors Chapa and Rao worked on the (bi)orthogonal relations to construct pattern adapted wavelet and have used it for target detection [4]. Akram Aldroubi, Patrice Abry, and Michael Unser used Projection based methods for generating pattern adapted wavelet [5]. Gupta et al. have suggested a method to construct wavelets that are matched to a given signal in the statistical sense [6]. Lifting scheme introduced by Wim Sweldens can be used to construct wavelets custom designed for certain applications [7].

Misiti et al. have used the method of least square optimization for generating pattern adapted wavelet. The principle for designing a new wavelet for CWT is to approximate a given pattern, using least squares optimization under constraints, leading to an admissible wavelet compatible for the pattern detection [8]. In our research work, we have used this method to generate adapted wavelet.

Detection of QRS complex in noisy ECG signal is not easy because of its varying nature. There are several methods developed to determine the accurate location of QRS peak using primary wavelets. Senhadji et al. has made a comparison on the ability of three different wavelets namely Daubechies, spline and Morlet to recognize and describe isolated cardiac beats [9]. Kadambe et al. applied DWT-based QRS detection algorithm. In their approach a specific spline wavelet, suitable for the analysis of QRS complexes, was designed and the scales were chosen adaptively based on the signal [10]. Li et al. suggested a method based on finding the modulus maxima larger than a threshold obtained from the pre-processing of preselected initial beats [11]. A quadratic spline wavelet with compact support was used in their analysis. The local maxima of the WT modulus at different scales were used to locate the sharp variation points of ECG signals.

Authors in [12] introduces a supervised learning algorithm which learns the optimal scales for each dataset using the annotations provided by physicians on a small training set. For each record, this method allows a specific set of non consecutive scales to be selected, based on the record's characteristics and along with a hard thresholding rule was used to label the R spikes. However pattern adapted wavelet for detection of R-peak in QRS complex provides a better performance in comparison with the above specified methods.

## III. RESEARCH METHODOLOGY

If we observe ECG waveforms we can find a set of variations of QRS complex. In order to study these variations data has been collected from www.physionet.org/physiobank. MIT-BIT Arrhythmia Database is used for performance comparison. Having analyzed the databank for many cases, typical five variation of QRS complex was selected. Using these five patterns five adapted wavelets was generated. CWT of these adapted wavelets with the practical QRS complexes provides maximum matching. This matching of QRS waveform with an adapted wavelet can maximize the coefficients in wavelet transform.

Pattern adapted wavelets are designed using least squares optimization method under certain constraints. The generated adapted wavelets were used in CWT analysis. The algorithm consists of two phases. The first phase of the algorithm decides the best matching wavelet for the QRS complex in the given ECG signal. For this purpose 250 points (approximately one cycle) of the ECG signal to be analyzed is extracted. This portion of the signal under consideration was transformed using CWT for each of the five pattern adapted wavelets. The maximum value of coefficients in each case will be compared. The adapted wavelet which gives maximum coefficients is selected for further analysis of ECG waveform.

In the second phase using the selected adapted wavelet, CWT of the given ECG signal is computed. From the coefficient matrix of CWT the position and scale of each peak is extracted. The peak values corresponding to P and T waves are removed to avoid false detection.

## IV. IMPLEMENTATION

From the physiobank, MIT-BIT Arrhythmia Database is used for selection of typical patterns. Five typical variations of QRS patterns are selected. A wavelet was adapted for each pattern using Matlab® Wavelet toolbox command 'pat2cwav', i.e. 'pattern to continuous wave'. The function gives an approximation to the given pattern in the interval [0 1] by least squares fitting a projection on the space of functions orthogonal to constants.

Fig. 1 through fig. 5 depicts the five adapted wavelets used for analysis. These adapted wavelets were used for the CWT analysis of ECG waveforms from the databank. A portion of ECG signal (approximately 250 points) is considered at the beginning. Using each wavelet, CWT is calculated for the signal under consideration. The adapted wavelet which gives maximum coefficient of CWT analysis is then considered for further analysis. In the second phase, by means of the best matched adapted wavelet the given ECG signal under goes a second time CWT analyzes. By further inspection of coefficient matrix of CWT R-peaks of ECG signal was extracted. One possibility of false detection is the peak of T-wave. After finding one QRS peak the searching for next QRS peak has to begin after an absolute refractory period. The important aspect is that it helps in preventing a T-wave being identified as an R-wave. The value used for absolute refractory period is 192milli seconds. Using these wavelets it was possible to find the QRS peaks accurately.

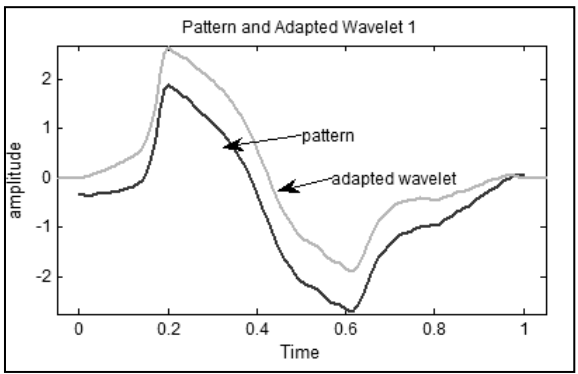

Figure 1. Adapted wavelet 1

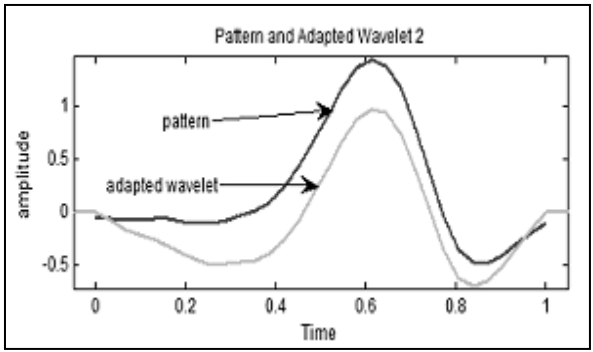

Figure 2. Adapted wavelet 2

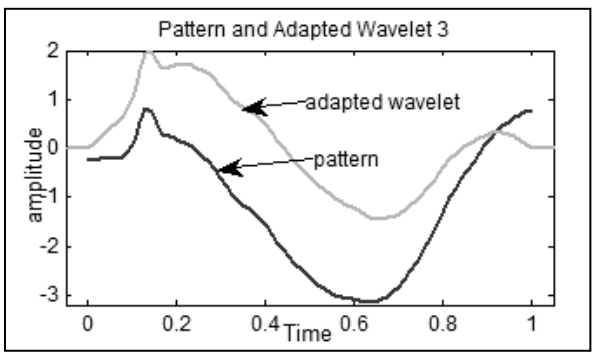

Figure 3. Adapted wavelet 3

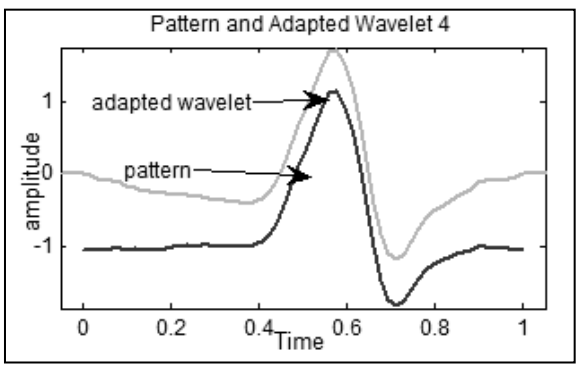

Figure 4. Adapted wavelet 4

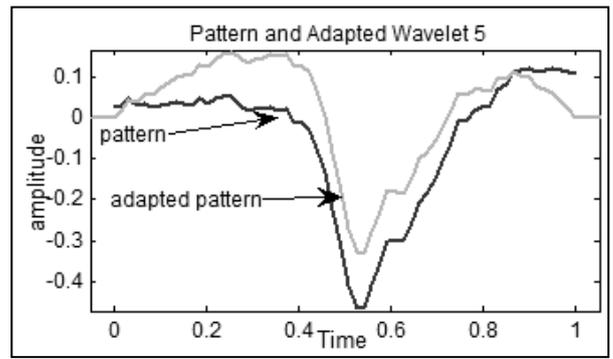

Figure 5. Adapted wavelet 5

## V. RESULT

Using adapted wavelet it was possible to find out R-peaks very accurately which is an important task in ECG analysis. Different ECG waveforms from the database have been tested and R-peak detection was accurate. Data from different ECG electrode has been tested and even the negative R- peaks were detected correctly. Even in the presence of baseline drift and high T-wave the detection was good. The adapted wavelets used with different waveforms are listed in table1. Fig. 6 to fig. 8 shows the R- peaks detected using adapted wavelets. R-peaks are represented using circles.

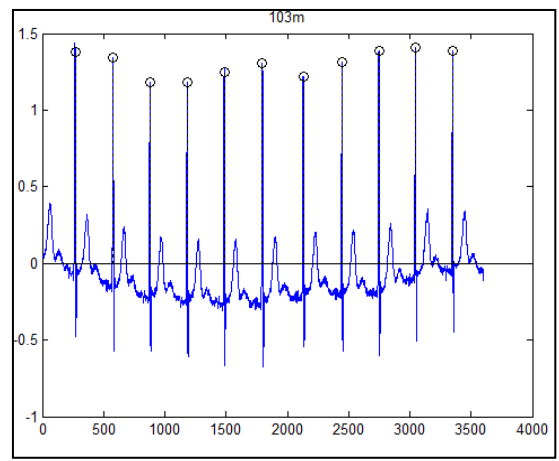

Figure 6. ECG waveform 103m analysed

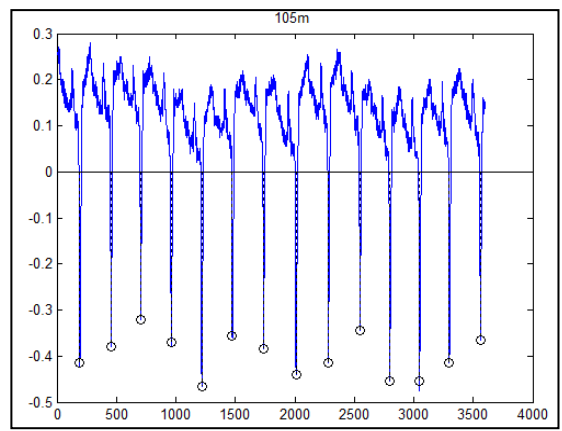

Figure 7. ECG waveform 105m analysed

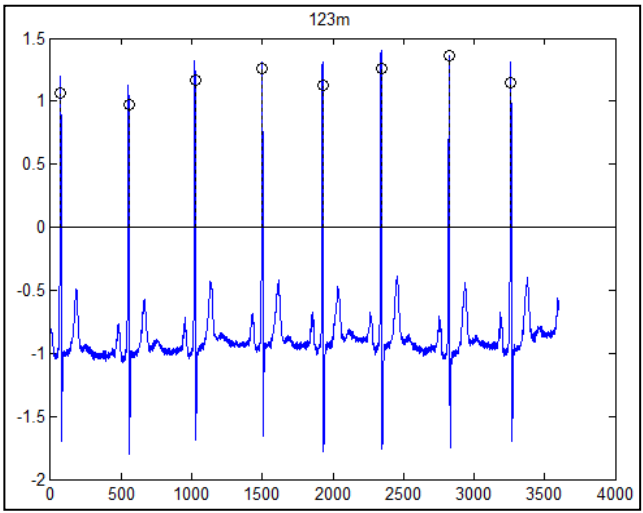

Figure 8. ECG waveform 123m analysed

TABLE I. ADAPTED WAVELET USED IN ANALYSIS

| Sl.No | Sample waveform | Adapted wavelet used |
|---|---|---|
| 1. | 102m V5 | ADW 3 |
| 2. | 105m MLII | ADW 5 |
| 3. | 103m MLII | ADW 2 |
| 4. | 107m V1 | ADW 1 |
| 5. | 123 V1 | ADW 4 |

## Conclusion

In this research work we have used secondary wavelets for the evaluation of CWT. This adaptive wavelet transform technique has successfully identified the R-peaks in most of the cases. Many researchers have worked with wavelet transforms as means of analyzing ECG signals. The methods used by them have been different and of varying effectiveness. All of them used primary wavelets for their analyzes and this paper shows a clear advancement on the efficacy of analyzing ECG signals with varational symptoms. The next step in this research will be to fine tune the algorithm to make it statistically successful across a wide variety of variations.

## Acknowledgement

The authors are grateful to Dr.Suma for her suggestions and comments on this paper.

## References

[1] Mallat, S. "A Wavelet Tour of Signal Processing. In: Wavelet Analysis and Its Applications," 2nd edn. IEEE Computer Society Press, San Diego 1999

[2] Paul S Addison "Wavelet transforms and the ECG:A Review," Cardio Digital Ltd, Elvingston Science Centre, East Lothian, EH33 1EH, UK.

[3] Jiapu pan and Willis J. Tompkins,"A Real-Time QRS Detection Algorithm," IEEE, IEEE Transactions on Biomedical Engineering, vol. BME-32, no. 3, March 1985

[4] Joseph O. Chapa and Raghuveer M. Rao "Algorithms for Designing Wavelets to Match a Specified Signal, " IEEE transactions on signal processing, vol. 48, no. 12, December 2000

[5] Akram Aldroubi, Patrice Abry, and Michael Unser "Construction of Biorthogonal Wavelets Starting from Any Two Multiresolutions," IEEE transactions on signal processing, vol. 46, no. 4, April 1998

[6] Gupta, A. Joshi, S.D. Prasad, S. "A new approach for estimation of statistically matched wavelet" IEEE Transactions on Signal Processing,vol.53 no.5 May 2005

[7] Wim Sweldens , Peter schroder "Building Your Own Wavelets at Home," Technical report Industrial Mathematics initiative. Department of mathematics, University of South Carolina 1995.

[8] Misiti, Y. Misiti, G. Oppenheim, J.M. Poggi, Hermes, "Les ondelettes et leurs applications," M, 2003.

[9] Senhadji L, Carrault G, Bellanger J J and Passariello G "Comparing wavelet transforms for recognizing cardiac patterns," IEEE Trans. Med. Biol. 13 167–73, 1995

[10] Kadambe S,Murray R and Boudreaux-Bartels G F "Wavelet transform-based QRS complex detector," IEEE Trans.Biomed. Eng. 46 838–48,1999

[11] Li C, Zheng C and Tai C "Detection of ECG characteristic points using wavelet transforms," IEEE Trans. Biomed.Eng. 42 21–8 1995.

[12] A. Fred, J. Filipe, and H. Gamboa "A Supervised Wavelet Transform Algorithm for R Spike Detection in Noisy ECGs," Biostec, CCIS 25, pp. 256–264, c Springer-Verlag Berlin Heidelberg 2008